%% file: main.tex
\documentclass[journal,twoside,web]{ieeecolor}

\usepackage{generic}
\usepackage{cite}
\usepackage{amsmath,amssymb,amsfonts}
\usepackage{algorithmic}
\usepackage{graphicx}
\usepackage{textcomp}
\usepackage{url}
\renewcommand{\baselinestretch}{0.982}

\usepackage{moreverb,url}
\usepackage[dvipsnames]{xcolor}
\usepackage{graphicx}
\usepackage{subfig}
\usepackage{algorithm}
\let\labelindent\relax
\usepackage{enumitem}
\usepackage[capitalise]{cleveref}

\usepackage{booktabs}

\newcommand{\highest}[1]{\textcolor{Maroon}{{#1}}}%
\newcommand{\bs}{\mathbf{s}}
\newcommand{\ba}{\mathbf{a}}

\input{math_commands.tex}

\def\BibTeX{{\rm B\kern-.05em{\sc i\kern-.025em b}\kern-.08em
    T\kern-.1667em\lower.7ex\hbox{E}\kern-.125emX}}
\begin{document}
\title{Robo-taxi Fleet Coordination at Scale via Reinforcement Learning}
\author{Luigi Tresca\authorrefmark{1}, Carolin Schmidt\authorrefmark{1}, James Harrison, Filipe Rodrigues, Gioele Zardini, Daniele Gammelli\authorrefmark{2} and Marco Pavone\authorrefmark{2}
\thanks{\authorrefmark{1} These authors contributed equally to this work.}
\thanks{\authorrefmark{2} These authors advised this work equally.}
\thanks{Manuscript submitted on April 8, 2025}
\thanks{Luigi Tresca is with the Department of Energy, Politecnico di Torino, Turin, Italy (e-mail: luigi.tresca@polito.it). }
\thanks{Carolin Schmidt and Filipe Rodrigues are with the Technical University of Denmark, Kongens Lyngby, Denmark (e-mail: {csasc, rodr}@dtu.dk).}
\thanks{James Harrison is with Google DeepMind, San Francisco, CA, USA (e-mail: jamesharrison@google.com).}
\thanks{Gioele Zardini is with the Laboratory for Information and Decision Systems, MIT, Cambridge, MA, USA (e-mail: gzardini@mit.edu).}
\thanks{Daniele Gammelli and Marco Pavone are with the Department of Aeronautics and Astronautics, Stanford University, Stanford, CA, USA (e-mail: {gammelli, pavone}@stanford.edu).}}

\maketitle

\begin{abstract}
Fleets of robo-taxis offering on-demand transportation services, commonly known as Autonomous Mobility-on-Demand (AMoD) systems, hold significant promise for societal benefits, such as reducing pollution, energy consumption, and urban congestion.
However, orchestrating these systems at scale remains a critical challenge, with existing coordination algorithms often failing to exploit the systems' full potential. 
This work introduces a novel decision-making framework that unites mathematical modeling with data-driven techniques.
In particular, we present the AMoD coordination problem through the lens of reinforcement learning and propose a graph network-based framework that exploits the main strengths of graph representation learning, reinforcement learning, and classical operations research tools.
Extensive evaluations across diverse simulation fidelities and scenarios demonstrate the flexibility of our approach, achieving superior system performance, computational efficiency, and generalizability compared to prior methods.
Finally, motivated by the need to democratize research efforts in this area, we release publicly available benchmarks, datasets, and simulators for network-level coordination alongside an open-source codebase designed to provide accessible simulation platforms and establish a standardized validation process for comparing methodologies. Code available at: \url{https://github.com/StanfordASL/RL4AMOD}
\end{abstract}

\begin{IEEEkeywords}
Fleet coordination, mobility-on-demand, network flow, reinforcement learning.
\end{IEEEkeywords}

\section{Introduction}
\label{sec:introduction}

\noindent More than half of the world's population now lives in urban areas, a figure projected to rise to $68\%$ by 2050~\cite{UNDepEconAff2021}.
Rapid urbanization is driving a surge in urban travel, intensifying the externalities associated with transportation, including greenhouse gas emissions, dependency on oil, and traffic congestion, to name a few.
Increasingly congested transportation systems are considered responsible for 30\% of emissions within urban environments and 50\% of health-related costs due to car-generated air pollution~\cite{OECD2014}. 
Furthermore, inefficiencies in traffic flow result in significant time loss and economic impact. In the US, congestion causes 8.8 billion hours of extra travel time, equivalent to more than a week of vacation for the average US commuter.
These challenges underscore the unsustainable nature of current transportation paradigms, centered around private car ownership.
Cities must, therefore, address the urgent need for mobility systems that balance growing transportation demands with environmental sustainability. 

One promising response to these challenges is the paradigm of \emph{robo-taxis}, shared mobility services that leverage autonomous vehicles (AVs) for on-demand urban travel, known as Autonomous Mobility-on-Demand (AMoD). Unlike private vehicles, robo-taxis enable service operators to centrally coordinate fleet operations, simultaneously managing passenger assignments and repositioning idle vehicles to mitigate imbalances.
By increasing vehicle utilization, AMoD systems can mitigate issues such as pollution, traffic congestion, and parking demand while simultaneously offering flexible and personal mobility options and reducing the capital costs associated with private vehicles~\cite{KuhnimhofEtAl2023}.

However, suboptimal coordination strategies can undermine the benefits of AMoD, leading to increased passenger wait times or traffic congestion caused by unnecessary movements of empty vehicles~\cite{OhLentzakisEtAl2021}.
Consequently, the large-scale deployment of robo-taxi services hinges on the ability to solve complex fleet coordination problems efficiently and at scale. 
From a control perspective, the coordination of robo-taxi fleets is a nonlinear, stochastic, and high-dimensional decision-making problem, posing significant challenges for conventional control techniques.


For instance, optimization-based methods often struggle with computational feasibility or depend on oversimplified assumptions about system dynamics and cost structures to remain tractable~\cite{hillier1967introduction}.
Similarly, data-driven approaches, often based on reinforcement learning (RL), have shown limited success in such contexts. 
Specifically, learning-based approaches are often sensitive to distribution shifts in unpredictable ways, whereas optimization-based policies are more readily characterized both in terms of robustness and out-of-distribution behavior~\cite{CobbeEtAl2019}.
Furthermore, data-driven methods often require substantial amounts of training data and struggle with the high-dimensional nature of the decision-making problem~\cite{ZhangEtAl2021, GammelliHarrisonEtAl2023}. 

In this work, we present a novel framework for coordinating robo-taxi fleets, integrating traditional optimization techniques from network control with state-of-the-art data-driven methods to address the challenges of nonlinear, stochastic, and high-dimensional decision-making.
To achieve this, we develop a learning-based framework that integrates the strengths of graph representation learning, RL, and classical operations research techniques\footnote{Although this paper focuses on AMoD systems, the proposed approach can be applied to MoD systems (without AVs) if there are compliant drivers~\cite{YangTsaoEtAl2021}.}.
Our approach starts with a graph representation of the transportation network, where nodes represent urban areas and edges denote connectivity between them~\cite{ZhangPavone2016}.
Leveraging this structure, we employ deep RL to learn a fleet coordination policy capable of optimizing vehicle movements to rebalance the system.
We argue that graph neural networks (GNNs) exhibit several desirable properties and propose an actor-critic formulation as a general approach to learning proactive, scalable, and transferable rebalancing policies.

\noindent\textbf{Contributions}. 
The contributions of this work are threefold.
%

First, we introduce a novel algorithmic framework for robo-taxi fleet control that integrates the strengths of optimization, reinforcement learning, and graph representation learning.
The optimization component provides a robust mathematical foundation for decision-making, offering both performance and safety guarantees.
Learning-based methods provide computational efficiency, adaptability to handle stochastic environments, and (delayed) reward maximization.
Finally, graph representation learning enables structured information propagation across networks with diverse topologies.
%
%

Second, we rigorously evaluate the proposed framework through a comprehensive set of experiments.
Unlike previous studies, our approach covers a broad spectrum of urban scenarios and leverages multiple simulation fidelities, ranging from coarse macroscopic to detailed mesoscopic simulations.
We also conduct an in-depth analysis of the transfer and generalization capabilities of deep RL approaches for robo-taxi coordination. 
Specifically, we assess the adaptability of learned policies in challenging contexts, including (i) inter-city generalization, (ii) network disturbances triggered by special events, and (iii) transferability to higher simulation fidelities. 
To further support and democratize research in this domain, we release an open-source codebase comprising benchmarks, datasets, pre-trained models, and simulators tailored for network-level coordination.
%
%

Third, we conduct detailed ablation studies on the architectural components and design choices within our framework.
This analysis covers the policy structure, neural network architecture,  learning signal, and their respective impacts on both system performance and the transferability of policies to unseen scenarios.

This work builds on our preliminary work~\cite{GammelliYangEtAl2021, GammelliYangEtAl2022, GammelliHarrisonEtAl2023, schmidt2024}, providing an extensive, large-scale evaluation of the proposed method using high-fidelity traffic simulations.

\section{Related Work}
\label{sec:relatedwork}
%
%
\noindent Our work lies at the intersection of network control, AMoD fleet coordination, and transferability of learning-based policies.
In this section, we discuss related literature and how existing approaches inform our work.
%

\noindent\textbf{Optimal and Learning-based Network Control.} \noindent Many economically critical real-world systems can be naturally formulated as control problems over graphs.
Effective system-level coordination is essential in diverse domains, including power generation~\cite{huneault1991survey}, transportation networks~\cite{WangSzetoEtAl2018, GammelliYangEtAl2021}, supply chain and distribution systems~\cite{bellamy2013network}, and telecommunication networks~\cite{popovskij2011control}.
These scenarios involve controlling the flow of products, vehicles, information, or other resources within network-structured environments.
A collection of highly effective solution strategies have been developed for these problems.
Some of the earliest applications of linear programming focused on network optimization~\cite{dantzig1982reminiscences, hillier1967introduction}
with multi-stage decision-making often addressed through time-expansion techniques~\cite{FordFulkerson1958}.
However, despite their broad applicability, these methods face significant scalability challenges.
Large-scale time-expanded networks can quickly become computationally intractable, ultimately limiting their practical utility in real-world scenarios.
More recently, data-driven approaches have emerged as an appealing paradigm to balance efficiency and optimality~\cite{Luong2019, Haydari2022its_survey}.
Despite their promise, such methods have seen limited real-world adoption, partly due to their sensitivity to distributional shifts and unpredictable performance in dynamic environments.
In this work, we introduce a hierarchical framework that combines traditional optimization techniques with advanced data-driven methods.
Our approach is designed to overcome the challenges of nonlinear, stochastic, and high-dimensional decision-making, bridging the gap between theory and practical application.

\noindent\textbf{AMoD Fleet Coordination.}
\noindent A particularly promising application of our hierarchical framework is the coordination of AMoD systems, which require real-time, large-scale vehicle coordination in complex and dynamic environments.
Managing such systems typically involves a trade-off between computational efficiency and optimality, with existing methods generally falling into one of three main categories~\cite{ZardiniLanzettiEtAl2021}.
Early AMoD coordination approaches often relied on rule-based heuristics~\cite{HylandMahmassani2018,Liu2019}.
While these methods are computationally efficient, they rarely achieve near-optimal performance.
In contrast, optimization-based techniques, commonly implemented using Model Predictive Control (MPC)~\cite{ZhangPavone2016, ZhangRossiEtAl2016b},
can guarantee optimal solutions under perfectly predictable conditions.
However, these methods often struggle with scalability in large networks and typically overlook the uncertainties inherent in real-world systems.
More recently, learning-based frameworks have emerged to address scalability challenges and manage demand uncertainties while maintaining near-optimal performance.
For instance, Mao et al.~\cite{MaoLiuEtAl2020} developed an actor-critic algorithm with an action space independent of fleet size, improving both scalability and flexibility.
Feng et al.~\cite{FengEtAt2021} proposed a centralized controller that sequentially assigns tasks to drivers, thus efficiently scaling with large fleets.
He et al.~\cite{HeEtAl2023} introduced a multi-agent RL approach for electric-AMoD systems under uncertainty, 
achieving robust performance through decentralized decision-making.

While these studies highlight the potential of learning-based methods in managing large systems, significant gaps remain.
Current approaches often fail to leverage the intrinsic graph structure of transportation networks when modeling decision policies, leading to suboptimal outcomes.
Furthermore, performance evaluations are frequently conducted in simplified traffic scenarios, limiting their applicability to real-world settings.
Our proposed solution addresses these limitations through a neural network architecture tailored to graph-structured data.
Furthermore, we validate our approach leveraging a multi-fidelity evaluation strategy, including extensive testing with a high-fidelity SUMO-based simulator~\cite{LopezEtAl2018sumo} calibrated with real-world data~\cite{Codeca2017}.
This holistic evaluation framework promotes robustness and practicality in diverse operational contexts.
\noindent\textbf{Policy Transferability.}
\noindent From a real-world perspective, a significant yet underexplored challenge in learning-based control of AMoD systems is the ability to \emph{transfer} control policies effectively across different urban environments.
Training a new controller from scratch for each city is both costly and time-consuming, underscoring the need for adaptable policies that maintain performance across varying traffic and demand distributions.
However, most existing research focuses on single-city scenarios~\cite{FluriRuchEtAl2019,HollerVuorioEtAl2019,SkordilisHouEtAl2021}.
Even studies addressing transferability primarily consider changes in demand patterns or congestion levels~\cite{SkordilisHouEtAl2021, LuoEtAl2023}, often neglecting performance in entirely different cities and omitting explicit transfer performance metrics in their training pipelines.
In contrast, our work directly integrates transferability objectives into AMoD policy design.
Through carefully crafted neural network architectures and strategic selection of learning signals, we aim to develop control strategies that deliver high performance across diverse conditions, including geographical variations (e.g., distinct network topologies), temporal fluctuations in traffic and demand (e.g., disruptions), and different levels of simulator fidelity.
Ultimately, our goal is to enable the efficient deployment of learned AMoD control policies across multiple urban contexts without the need for extensive retraining, significantly reducing operational costs. 


\section{Background}
\label{sec:background}
\noindent In this section, we first outline the RL problem (\cref{subsec:rl_problem}), then introduce two traffic simulation paradigms, the macroscopic in \cref{subsubsec:macro_sim} and the mesoscopic in \cref{subsubsec:meso_sim}, that underlie our AMoD system evaluations.

\subsection{The Reinforcement Learning Problem}
\label{subsec:rl_problem}
\noindent In RL, our goal is to learn to control a dynamic system through experience~\cite{SuttonBarto1998}. 
Specifically, we define a fully observed Markov Decision Process (MDP)~$\mathcal{M} = \left(\mathcal{S}, \mathcal{A}, P, d_0, r, \gamma \right)$, where~$\mathcal{S}$ represents a set of states~$\bs \in \mathcal{S}$, which may be discrete or continuous, $\mathcal{A}$ is a set of possible actions $\ba \in \mathcal{A}$, also discrete or continuous. 
The system’s dynamics are described by a conditional probability distribution $P(\bs_{t+1} | \bs_t, \ba_t)$, while the initial state distribution is defined by $d_0(\bs_0)$, the reward function by $r : \mathcal{S} \times \mathcal{A} \rightarrow \mathbb{R}$, and the scalar discount factor by $\gamma \in (0,1]$. 
At a high level, the objective is to learn a policy $\pi(\ba_t | \bs_t)$, which is a distribution over actions given states. 
The learning process aims at maximizing the expected cumulative reward through interactions with the MDP $\mathcal{M}$.
The term \emph{trajectory} refers to a sequence of states and actions over a horizon $H$, represented as $\tau = (\bs_0, \ba_0, \ldots , \bs_H, \ba_H)$, where $H$ may be infinite.
Using this terminology, we can derive the \emph{trajectory distribution} $p_{\pi}$ for a given policy $\pi$ as $p_{\pi}(\tau) = d_0(\bs_0) \prod_{t=0}^H \pi(\ba_t | \bs_t) P(\bs_{t+1} | \bs_t, \ba_t)$.
The RL objective $J(\pi)$ can then be expressed by the expected cumulative reward over $p_{\pi}$: $J(\pi) = \mathbb{E}_{\tau \sim p_{\pi}(\tau)} \left[\sum_{t=0}^{H} \gamma^t r(\bs_t, \ba_t) \right]$.


In most RL algorithms, the learning process relies on data generated through interactions between an agent and MDP~$\mathcal{M}$, i.e., the environment, using a \emph{behavior policy} that may or may not align with the target policy~$\pi(\ba | \bs)$.
Typically, this involves observing a state $\bs_t$, selecting an action $\ba_t$, and receiving the next state $\bs_{t+1}$ along with a reward $r_t = r(\bs_t, \ba_t)$. 
Learning methods then leverage these transitions~$(\bs_t, \ba_t, \bs_{t+1}, r_t)$ to iteratively improve the policy. 
In this work, we consider a broad spectrum of learning paradigms—including online RL, meta-RL, offline RL, and behavior cloning—each characterized by distinct approaches to data collection and use during training.

\subsection{Traffic Simulators}\label{subsec:traffic_sim}
\noindent Traffic simulation models vary in detail and computational complexity, from macroscopic simulators that analyze traffic at an aggregate level to microscopic simulators modeling individual vehicles. 
Mesoscopic simulators lie in between, balancing microscopic accuracy with macroscopic efficiency. 
Since we focus on network-level characteristics rather than individual vehicle behaviour, we focus on macroscopic and mesoscopic approaches for modeling AMoD systems. 

\noindent\textbf{Macroscopic Approach.}
\label{subsubsec:macro_sim}
The macroscopic approach models traffic at a high level of abstraction, focusing on aggregate variables such as density, flow, and velocity over road networks. An AMoD system is represented as a station-based network where the city is clustered into regions \(\mathcal{V}\), used as the nodes of a directed graph \(\mathcal{G} = (\mathcal{V}, \mathcal{E})\). Each edge \((i,j) \in \mathcal{E}\) represents travel from region \(i\) to region \(j\), with \(\tau_{ij}\) denoting the travel time along the shortest path connecting their centroids. The aggregate flow on edge \((i,j)\) at time \(t\) is denoted by \(f_{ij}^t\), which is further split into passenger flow \(\paxflow_{ij}^{t}\) and rebalancing flow \(\rebflow_{ij}^{t}\).
To capture changes in the system state, the model tracks the number of idle vehicles \(M_{i}^t\) in each region \(i\). The idle vehicle dynamics is computed by balancing arrivals (i.e., vehicles that have completed inter-region travel) against departures for new passenger trips or rebalancing actions, formally given by 
\(\dot{M}_{i}^{t} = \sum_{j \in \mathcal{V}}\bigl(\paxflow_{ji}^{t - \tau_{ji}} + \rebflow_{ji}^{t - \tau_{ji}}\bigr) - \sum_{j \in \mathcal{V}}\bigl(\paxflow_{ij}^{t} + \rebflow_{ij}^{t}\bigr)\).
Likewise, the inter-region flow evolves according to 
\(\dot{f}_{ij}^{t} = \bigl(\paxflow_{ji}^{t} + \rebflow_{ji}^{t}\bigr) - \bigl(\paxflow_{ji}^{t - \tau_{ji}} + \rebflow_{ji}^{t - \tau_{ji}}\bigr)\).
Additionally, the framework enforces 
demand feasibility, 
\(\paxflow_{ij}^{t} \leq \demand_{ij}^{t}\), 
flow conservation, 
\(\sum_{i\in \mathcal{V}}\bigl(\paxflow_{ij}^{t - \tau_{ij}} + \rebflow_{ij}^{t - \tau_{ij}}\bigr) = \sum_{k\in \mathcal{V}}\bigl(\paxflow_{jk}^{t} + \rebflow_{jk}^{t}\bigr)\), 
and non-negativity constraints, 
\(\paxflow_{ij}^{t} \ge 0\), \(\rebflow_{ij}^{t} \ge 0\). 
This compact framework enables efficient analysis of system-wide AMoD dynamics while abstracting from individual vehicle behaviors.


\noindent\textbf{Mesoscopic Approach.}
\label{subsubsec:meso_sim}
The mesoscopic framework offers a hybrid approach to traffic simulation, bridging the gap between microscopic and macroscopic models by representing individual vehicle behaviors at an aggregated level and approximating vehicle movements with queuing theory. 
This allows for spatial aggregation of car flows, significantly reducing computational demands while still capturing key congestion dynamics.
Here, each network link is divided into queues of length \( L^k \). Each queue is characterized by the number of vehicles present, $n_k$, and a free-flow speed, $v_{\text{max}}$, which is the maximum allowable speed in the absence of congestion.
Additionally, a congestion threshold $n_{\text{jam}}$ is defined uniformly for all queues in the network to differentiate between congested and not-congested conditions.
Vehicle movement within the $k$th queue is modeled by defining travel time across the queue. 
In particular, under free-flow conditions, a vehicle $\nu$ spends a minimum time in the queue given by the free-flow travel time $t_{\text{ff}}^{\nu} = L^i/v^i_{\text{max}}$. In contrast, in congested conditions, the travel time for the $\nu$th vehicle in the $k$th queue is constrained by the preceding vehicle’s travel time as  $t_{\text{tt}}^{\nu} \geq t_{\text{tt}}^{\nu-1} + H_s^i(n^i, n^i_{\text{jam}})$.  In particular, the time headway $H_s^i$ represents the minimum time interval required between consecutive vehicles on the same lane to ensure safe driving conditions. Its value is adjusted based on the congestion state of the current and downstream queues, where $n^i_{\text{jam}}$ is a calibration parameter to model realistic congestion behavior.
Mesoscopic frameworks can be applied to real urban road networks, where intersections are modeled as uncontrolled, priority-based, or signalized nodes. 
The approach effectively captures network congestion and travel time dynamics while maintaining computational efficiency. 
We use both macroscopic and mesoscopic traffic models to investigate how simulator fidelity impacts policy performance and validate it in an environment that closely approximates real-world conditions.



\section{Robo-taxi Coordination as a Network Flow Problem}
\label{sec:problem_formulation}
\noindent In this section, we introduce the notation and terminology characterizing the robo-taxi fleet coordination problem and its formulation as a network flow problem.

\subsection{Robo-taxi Fleet Coordination}
\noindent An AMoD operator coordinates $\numvehicles$ fully AVs that provide on-demand mobility services within a given transportation network, represented as a graph $\graph = \left(\nodes, \edges\right)$, where $\nodes$ corresponds to the set of spatially aggregated stations (e.g., pickup or dropoff locations) and $\edges$ represents the set of links $\edge_{ij}$ connecting all adjacent stations $i, j$.
We denote the total number of stations as $\numregions = \lvert \nodes \rvert$.
The time horizon is divided into discrete intervals $\horizon=\{1,2,\cdots, T\}$, each with a duration of $\Delta T$. 
Vehicles traveling between station $i \in \nodes$ and station $j \neq i \in \nodes$ are controlled by the operator to follow the shortest path, with a travel time of $\ttime_{ij}^t \in \mathbb{Z}_+$ time steps and a travel cost of $\tcost_{ij}^t$.
Travel times are assumed to be dependent on congestion levels according to traffic flow models\footnote{The flow model is defined by the simulator fidelity (see \cref{subsec:traffic_sim}).}.
Passengers make trip requests at each time step. 
The requests with origin-destination pair $(i, j) \in \nodes \times \nodes$ received at $t$ are defined by demand $\demand_{ij}^t$ and price $\price_{ij}^t$\footnote{We assume passenger demand and price to be independent of the control of the AMoD fleet. 
This assumption can be relaxed by training the proposed RL model in an environment incorporating demand modeling and surge prices.}.
We further assume that passengers not matched with an AV within a maximum time $\maxwait \in \mathbb{Z}_+$ will leave the system. 
The objective of the operator is to dynamically coordinate the fleet to (i) match available AVs with passenger requests and bring them to their destination and (ii) control the idle AVs to either stay in their current station or be rebalanced to other stations to optimize performance. 
\begin{figure*}[tbh]
    \centering
    \includegraphics[width=\textwidth]{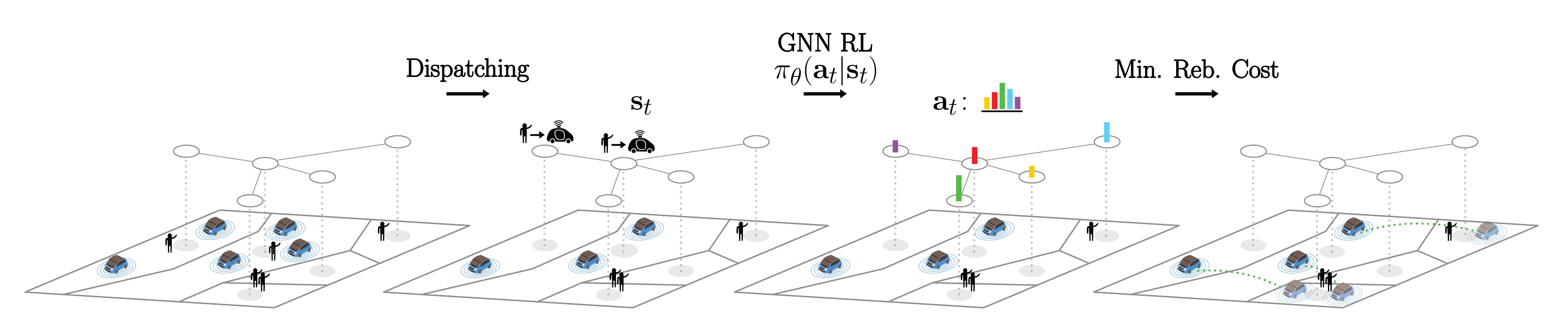}
    \caption{Illustration of the proposed hierarchical decomposition for AMoD fleet coordination. Given the current distribution of idle vehicles and customer transportation requests, the decomposition entails: (1) assigning idle vehicles to trip requests (i.e., $\paxflow_{ij}^t$) by solving a convex dispatching problem; (2) determining a desired future allocation of vehicles across regions (i.e., $\hat\st^t)$ via RL; and (3) converting $\hat\st^t$ into actionable rebalancing trips (i.e., $\rebflow_{ij}^t$) while minimizing the overall rebalancing cost.}
    \label{fig:3-step-framework}
\end{figure*}
\subsection{Network Flow Problem Formulation}
\noindent The AMoD control problem is naturally posed as a network flow problem.
We denote $\paxflow_{ij}^t \in \mathbb{N}$ as the passenger flow, i.e., the number of matched passengers traveling from station $i \in \nodes$ to station $j \in \nodes$ at $t$, and $\rebflow_{ij}^t \in \mathbb{N}$ as the rebalancing flow, i.e., the number of vehicles rebalancing from station $i \in \nodes$ to station $j \in \nodes$ at $t$.
From the operator's perspective, the objective is to maximize profit over a time horizon $T$:
\begin{subequations}
\begin{align}
    & \max_{\{(\paxflow_{ij}^t, \rebflow_{ij}^t)\}_{i,j\in\nodes, t \in \horizon}} \quad  \sum_{t=1}^{T} \sum_{i,j \in \nodes} [ ( \price_{ij}^{t} - \tcost_{ij}^{t})\paxflow_{ij}^{t} - \tcost_{ij}^{t} \rebflow_{ij}^{t}] \label{eq:global_obj} \\
    & \mathrm{s.\ t.} \quad  \paxflow_{ij}^{t} \leq \demand_{ij}^{t} \quad \forall i,j\in\mathcal{V},  t\in\mathcal{T} \label{eq:global_c1}\\
    & \sum_{i\in\nodes}  (\paxflow_{ij}^{t - \ttime_{ij}} + \rebflow_{ij}^{t - \ttime_{ij}})  = \sum_{ k\in\nodes} (\paxflow_{jk}^{t} + \rebflow_{jk}^{t})  \; \forall t\in\mathcal{T}, j \in \nodes \label{eq:global_c2}\\
    & \paxflow_{ii}^{0} = \paxflow^{\text{init}}_{i} \;,\; \rebflow_{ii}^{0} = 0 \quad \forall i\in\nodes \label{eq:global_c4}\\
    & \paxflow_{ij}^{t} \geq 0 \;,\; \rebflow_{ij}^{t} \geq 0, \quad \forall i\in\nodes, t\in\mathcal{T}
    \label{eq:global_c5}
\end{align}
\end{subequations}
where the objective term \cref{eq:global_obj} represents the total profit, \cref{eq:global_c1} ensures that passenger flow does not exceed demand, \cref{eq:global_c2} enforces flow conservation, and \cref{eq:global_c4} and \cref{eq:global_c5} set the initial conditions and ensure non-negativity of the decision variables, respectively.
Given $\lvert \edges \rvert = \mathcal{O}\left(\nodes^2\right)$, the optimization problem defined in \cref{eq:global_obj} involves $\mathcal{O}\left(\nodes^2 \times T\right)$ decision variables. 
The rapid growth of the problem’s complexity with increasing spatial resolution and planning horizon $T$ leads to significant computational challenges for real-time applications~\cite{SinghalGammelliEtAl2024}.
Moreover, to reduce complexity, Equations (\ref{eq:global_obj})-(\ref{eq:global_c5}) do not model sources of volatility inherent to real-world systems, such as uncertainty in demand patterns and travel times.
In this work, we aim to reduce the complexity of Problem 
 (\ref{eq:global_obj})-(\ref{eq:global_c5}) to enable real-time control while maintaining the performance guarantees of mathematical optimization and leveraging the ability of learning-based methods to adapt the control policy to unmodeled stochasticities.

\section{Graph RL for AMoD Fleet Control}
\label{sec:amod_problem}
\noindent In this section, we systematically introduce the key elements of the proposed approach. 
We begin by introducing a hierarchical decomposition of Problem (\ref{eq:global_obj})–(\ref{eq:global_c5}), which is designed to leverage the complementary strengths of optimization-based and learning-based control (\cref{subsec:decomposition}).
Next, we discuss the application of GNNs for addressing network control problems (\cref{subsec:grl}) and elaborate on the design of the learning signal used to train the RL policy (\cref{subsec:learning_signal}).

\subsection{Hierarchical Policy Decomposition}
\label{subsec:decomposition}
\noindent The previous section introduced the AMoD control problem via network optimization. 
From an MDP perspective, the goal is to find a policy $\pflatopt \in \pflatspace$ (where $\pflatspace$ is the space of realizable Markovian policies) that maximizes the long-term objective in \cref{eq:global_obj} while satisfying constraints (\ref{eq:global_c1})–(\ref{eq:global_c5}).
This formulation entails specifying a distribution over all feasible flow actions (i.e., a policy), which may be an extremely large and dynamic space.
As illustrated in \cref{fig:3-step-framework}, we consider a hierarchical decomposition, where, at each time $t$, we compute an action $\ac^t = \{(\paxflow_{ij}^t, \rebflow_{ij}^t)\}_{i,j \in \nodes}$ by replacing a single policy that maps from state to actions ($\st^t \rightarrow \ac^t$) into a composition of policies that maps states to desired states to actions ($\st^t \rightarrow \hat{\st}^t \rightarrow \ac^t$), resulting in a three-step framework:

\smallskip
\noindent\textbf{1. Dispatching $(\paxflow_{ij}^t)$}:
The first step entails assigning vehicles to customers and obtaining passenger flows $\{\paxflow_{ij}^t\}_{i,j \in \nodes}$:
\begin{subequations}
\begin{align}
    \max_{\{\paxflow_{ij}^t\}_{i,j\in\nodes}} \quad& \sum_{i,j\in\nodes}\paxflow_{ij}^t(\price_{ij}^t- \tcost_{ij}^t) \label{eq:matching_obj}\\
    \rm{s.t.} \,\,\,\,\, \quad& 0\leq \paxflow_{ij}^t \leq \demand_{ij}^t, ~i,j\in\nodes, \label{eq:matching_con1}\\
    & \sum_{j \in \nodes} x_{ij}^t \leq \numvehicles_{i}^t, ~i\in\nodes, \label{eq:matching_con2}
\end{align}\label{eq:matching}
\end{subequations}
where the objective function (\ref{eq:matching_obj}) represents the profit of passenger assignment calculated as the difference between revenue and cost, the constraint (\ref{eq:matching_con1}) ensures that the passenger flow is non-negative and upper-bounded by the demand,  and (\ref{eq:matching_con2}) enforces that the total passenger flow does not exceed the number of idle vehicles $\numvehicles_i^t$ at station $i$ at time step $t$. 
Notice that the constraint matrix of the assignment problem (\ref{eq:matching}) is totally unimodular~\cite{Nemhauser1988}, hence, the resulting passenger flows are integral as long as the demand is integral. 
At a high level, given the current state of the system, the linear program (LP) defined in Step 1 optimally assigns available AVs to customer requests with the objective of maximizing operator profit.

\smallskip
\noindent\textbf{2. Desired Vehicle Distribution $(\hat\st^t)$}:
Once available AVs have been assigned to customer requests, Steps 2 and 3 jointly compute rebalancing actions that strategically influence the future distribution of AVs to improve overall system performance.
Specifically, the second step entails computing a \textit{desired next state} as an intermediate action representation.
We are interested in defining a desired next state $\hat \st^{t+1}$ that is ideally (i) lower dimensional compared to $\ac^t = \{(\paxflow_{ij}^t, \rebflow_{ij}^t)\}_{i,j \in \nodes}$, (ii) able to capture relevant aspects for control, and (iii) as-robust-as-possible to domain shifts.
We achieve this by avoiding the direct parametrization of \textit{per-edge} desired flow values and computing \textit{per-node} desired availabilities. 
Concretely, given $\numvehicles$ available AVs in the network, we define $\hat \st^{t+1} = \{\hat \inventory_i^{t+1}\}_{i \in \nodes}, \sum_i \hat \inventory_i^{t+1} = \numvehicles$ as a desired per-node number of AVs.
We do so by first determining $\tilde{\inventory}_{i}^{t+1} = \{\tilde{\inventory}_{i}^{t+1}\}_{i \in \nodes}$, where $\tilde{\inventory}_{i}^{t+1} \in [0,1]$ defines the desired percentage of currently available vehicles to be located in region $i$ at time step $t$, and $\sum_{i \in \nodes} \tilde{\inventory}_{i}^{t+1} = 1$. 
We then use this to compute $\hat{\inventory}_i^{t+1}=\lfloor \tilde{\inventory}_{i}^{t+1} \cdot M\rfloor$ as the actual number of vehicles.
We achieve this by defining the intermediate policy as a Dirichlet distribution over nodes, i.e., $\pi(\hat \st^{t+1} | \st^t) =  \tilde{\inventory}^{t+1} \cdot M, \tilde{\inventory}^{t+1} \sim \text{Dir}(\tilde{\inventory}^{t+1} | \st^t)$.
Crucially, the representation of the desired next state via $\hat \inventory_i$ (i) is lower-dimensional as it only acts over nodes in the graph 
(ii) uses a meaningful aggregated quantity to control flows, and (iii) is scale-invariant by construction, as it acts on \textit{ratios} as opposed to raw flow quantities.
Ultimately, Step 2 receives as input the state of the system---after dispatching---and computes a desired future allocation of vehicles across regions in the transportation network.

\smallskip
\noindent\textbf{3. Minimum Cost Flow $(\rebflow_{ij}^t)$}:
Lastly, the third step entails transforming the desired next state from Step 2 into actionable rebalancing flows $\{\rebflow_{ij}^t\}_{i,j \in \nodes}$ by solving the following LP:
\begin{subequations}
\begin{align}
    \min_{\{\rebflow_{ij}^t\}_{i, j\in \nodes}} \quad& \sum_{i\neq j\in \nodes}\tcost_{ij}^t \rebflow_{ij}^t \label{eq:reb_obj}\\
    \rm{s.t.}  \quad \quad \quad \,
    & \sum_{j\neq i}(\rebflow_{ji}^t - \rebflow_{ij}^t) + M_i^t \geq \hat{\inventory}_i^t , ~i\in\nodes,\label{eq:reb_con1} \\
    \quad& \sum_{j\neq i} \rebflow_{ij}^t \leq M_i^t,~i\in\nodes,\label{eq:reb_con2}
\end{align}\label{eq:reb}
\end{subequations}
where the objective (\ref{eq:reb_obj}) represents the rebalancing cost, constraint (\ref{eq:reb_con1})  ensures that the \textit{resulting} number of vehicles is close to the \textit{desired} number of vehicles, and constraint (\ref{eq:reb_con2}) ensures that the total rebalancing flow from a region is upper-bounded by the number of idle vehicles in that region. 

\noindent\textbf{Overall Framework.}
The proposed three-step framework yields a policy that integrates learning-based (Step 2) and optimization-based (Steps 1 and 3) components.
This hierarchical decomposition is deliberately designed to exhibit three appealing properties for network control problems:

\noindent\textit{Scalability and Inference Time}. The computational complexity of Problem (\ref{eq:global_obj})–(\ref{eq:global_c5}) scales significantly with the planning horizon $T$. 
Instead of explicitly modeling future flow variables via time-expansion techniques, our approach decomposes the problem temporally: Steps 1–3 compute flow actions only for the current timestep. 
Long-horizon planning is handled implicitly through the RL policy, which is trained to optimize expected future rewards. This design enables fast inference, as fleet coordination reduces to a forward pass through a neural network (Step 2) and the solution of a convex optimization problem (Step 1 and 3), achieving scalable, real-time performance even in large transportation networks.

\noindent\textit{Non-linear and Stochastic Elements}. Traditional approaches that formulate fleet coordination as a network flow problem often omit stochastic elements---such as volatility in future travel times and customer requests---to maintain computational tractability. In contrast, our hierarchical decomposition purposefully applies exact optimization where it is most effective (i.e., short-term decision-making) while delegating the handling of long-term nonlinear and stochastic effects to the learned policy. This enables the framework to retain computational efficiency while adapting to uncertainties.

\noindent\textit{Handling of Constraints}. Learning-based methods for network control often struggle to enforce hard constraints, particularly in dynamic operational environments where system requirements may change frequently (e.g., new geographically localized regulations). By integrating learning-based and optimization-based components, our approach ensures constraint satisfaction by leveraging optimization to project the potentially infeasible outputs of the RL policy onto the space of feasible solutions. This not only allows the system to adapt to new constraints introduced after training but also provides the operator with direct control over policy outputs.


\subsection{Graph Representation Learning for Network Control}
\label{subsec:grl}
\noindent One of the key reasons for the success of deep neural networks lies in their ability to leverage statistical properties of the data through learnable hierarchical processing.
In the context of network optimization problems, we argue that Graph Neural Networks (GNNs) are a natural and effective modeling choice.
The GNN framework provides a versatile approach for relational reasoning over graph-structured data through GNN \emph{blocks}. 
These blocks perform computations on an input graph and return an updated output graph, implementing graph-to-graph operations. 
Specifically, GNNs are well-suited to our problem setting for three key reasons.

First, \emph{permutation invariance}.
Consider a transportation network represented as a graph with $\numregions$ nodes, where each node $i$ is associated with $d$-dimensional attributes $\bx_i \in \mathbb{R}^d$.
A computation is permutation-invariant if its output remains unchanged regardless of the node ordering.
Conversely, a non-permutation-invariant computation would treat different orderings as distinct, requiring an exponential number of training examples to generalize effectively.
In this work, we argue that permutation-invariant models, such as GNNs, align naturally with the dominant modeling paradigms in transportation engineering.
This stems from the observation that urban areas, akin to graph nodes, lack a natural ordering.
Instead, transportation networks are best characterized by node and edge attributes---such as regional demand or inter-region travel time---rather than by an arbitrary indexing of locations.

Second, \emph{locality of the operator}.
GNNs typically employ local parametric filters (e.g., convolution operators), allowing the same neural network to be applied to graphs of varying sizes and topologies.
This property is crucial for real-world network control problems, where the system may be dynamic or frequently reconfigured.
By leveraging locality, the same policy can generalize across different instances of the problem. 

Lastly, \emph{alignment with network optimization algorithms}.
While deep neural networks possess the \emph{capacity} to approximate a broad range of functions, not all architectures can efficiently \emph{learn} these functions.
Intuitively, a neural network is more likely to generalize well if it is structurally aligned with the target function.
A classical example is the relationship between MLPs and CNNs in computer vision---while MLPs are universal approximators, they struggle to achieve competitive performance on vision tasks due to a lack of built-in spatial priors.
As demonstrated in~\cite{MaXuEtAl2020}, GNNs naturally align with many network optimization algorithms, making them well-suited for learning effective control policies.

Motivated by these properties, our approach leverages GNNs to compute the desired vehicle allocation policy (i.e., $\pi(\hat{\st}^t \lvert \st^t)$) described in Step 2 of our policy decomposition. 
%
%
\subsection{Learning Signal}
\label{subsec:learning_signal}
\noindent The hierarchical policy described in this section combines optimization-based methods with learning.
This learning process can take place in an online setting, either within a single-task \cite{SuttonBarto1998} or multi-task \cite{FinnAbbeelEtAl2017} framework, or it can be based on offline data, leveraging either behavior cloning \cite{HusseinEtAl2018} or offline RL \cite{LevineEtAl2020} techniques.
In the following, we discuss the role of these paradigms within the scope of AMoD control. 

\smallskip
\noindent\textbf{Online RL.}
Online policy learning is the most common RL paradigm, where policies are learned through interaction with an environment.
In the context of AMoD coordination, this typically requires access to a calibrated traffic simulator.
However, to the best of the authors’ knowledge, there has been no extensive study on how simulator fidelity affects learning performance, with most studies relying on lower-fidelity macroscopic simulations.
We consider the actor-critic methods A2C \cite{SuttonBarto1998} and SAC \cite{Haarnoja2018} for online learning. 


\smallskip
\noindent\textbf{Meta-RL.}
Current learning-based approaches for controlling AMoD systems are limited to the \textit{single-city} scenario, where the service operator is allowed to make an unlimited number of operational decisions within the same transportation system \cite{FluriRuchEtAl2019,HollerVuorioEtAl2019,SkordilisHouEtAl2021}.
However, real-world operators cannot afford to fully retrain AMoD controllers for every city they operate in. 
We propose to formalize the \textit{multi-city} AMoD control problem from a meta-RL perspective, enabling agents to leverage prior experience for rapid adaptation to new cities.
Specifically, we define each city as a new task and propose meta-RL through recurrence \cite{duan2016rl} as a framework for learning policies that achieve strong performance in new cities with limited online operational decisions.

\smallskip

\noindent\textbf{Offline RL.}
At its core, any \textit{online} learning algorithm assumes access to a reliable traffic simulator.
However, for many real-world instances, reliable simulation may not be available, computationally expensive to interact with, or exposed to sim-to-real distribution shifts, while historical operational data is often abundant.
Thus, offline RL represents a promising direction to overcome these challenges by learning solely from real, pre-collected data, avoiding expensive real-world exploration. 
In this work, we consider two offline RL algorithms, conservative Q-learning (CQL) \cite{Kumar2020} and implicit Q-learning (IQL) \cite{KostrikovEtAl2022}. 

\smallskip
\noindent\textbf{Behavior Cloning (BC).}
Similar to offline RL, BC relies solely on pre-collected data by training a policy to imitate expert demonstrations through supervised learning.
While the performance of BC-learned policies is inherently limited by the quality of the demonstrations, BC provides a stable learning signal. 
This makes BC particularly useful for policy learning when expert demonstrations are available or, when this is not the case, as a pre-training phase before further refinement through other learning techniques. 

\smallskip
\noindent In this work, we explore the benefits and limitations of these learning paradigms for AMoD control and demonstrate how our hierarchical decomposition is amenable to them.

\section{Experiments}
\label{sec:experiments}
\noindent The experiments are structured to systematically evaluate the effectiveness of our graph RL approach across simulator fidelity levels. We begin by outlining the characteristics of the macroscopic and mesoscopic scenarios, followed by a description of our benchmarks. 
Our evaluation begins within a macroscopic simulation in \cref{subsec:macro_results} to examine key design choices in our framework. Specifically, we assess the effectiveness of using GNNs as neural architecture for decision-making over networks, the computational benefits of our hierarchical policy decomposition, and the impact of the choice of learning signal on transferability, adaptability, and sample efficiency. Finally, we conduct an in-depth analysis at the mesoscopic fidelity level in \cref{subsec:meso_results}, investigating the influence of simulation fidelity on policy performance and the generalization capabilities of the learned policy across scenarios and fidelity levels.
To the best of the authors' knowledge, this work represents the first multi-fidelity evaluation of RL-based control policies for robo-taxi fleet coordination.


\subsection{Scenarios}
\label{subsec:city_scenario}
\noindent We categorize the evaluation scenarios according to fidelity:  

\noindent\textbf{Macroscopic Scenarios.} 
We study hypothetical deployments of taxi-like systems to meet peak-time demand in different case studies from cities around the world. Specifically, we focus on traffic scenarios in New York (USA), Shenzhen (China), and San Francisco (USA). 
The case studies use publicly available trip record datasets, which are converted into demand, travel times, and trip prices between stations, available in our code repository. We assume travel times are given and independent of operator actions, which is appropriate for cities where the fleet size is small relative to the total vehicle population, making the impact of traffic congestion negligible. 
Further, we assume that passenger arrival processes are independent for each OD-pair and are modeled using a time-dependent Poisson process \cite{Daganzo1978}.

\noindent\textbf{SUMO-based Mesoscopic Scenario.}
We develop an AMoD system based on a mesoscopic traffic model (see \ref{subsubsec:meso_sim} for details) within the SUMO simulator~\cite{LopezEtAl2018sumo}: a microscopic and continuous traffic simulation. This scenario incorporates exogenous traffic and road infrastructure elements such as signs, lanes, and traffic lights. These elements introduce a dynamic response influenced by operator actions and policy-induced congestion, affecting the travel time. 
The policy performance is assessed in the city of Luxembourg using a calibrated traffic scenario based on 24 hours of real traffic data. Details on the calibration of the scenario can be found in~\cite{Codeca2017}.
The original network has been spatially aggregated into different regions. 
Trip prices were set according to the pricing scheme of Luxembourg\footnote{Available at: https://www.bettertaxi.com/taxi-fare-calculator/luxemburg/}, while the costs were calibrated using publicly available data from Lyft\footnote{Cost structure dataset available at: https://www.lyft.com/pricing/SFO}, ensuring realistic representation of the cost structures for the AMoD system.

\subsection{Baseline Control Strategies}
\label{subsec:benchmarks}

\noindent In our experiments, we compare the results of our proposed graph-RL-based policy with several baselines, spanning from pure optimization methods to rule-based strategies: 

\noindent\textbf{No Rebalancing}: the system performance is measured without any rebalancing activities.

\noindent\textbf{Heuristics}:
\begin{itemize}[leftmargin=*]
    
    \item \emph{Equally distributed}: at each decision step, the heuristic aims to distribute idle vehicles evenly across the network.
    
    \item \emph{Plus-one}~\cite{ruch2020the+}: 
    For every vehicle that departs from a region to fulfill a passenger trip, another vehicle is rebalanced back to that same region.
\end{itemize}

\noindent\textbf{Optimization-based}:
\begin{itemize}[leftmargin=*]
    \item \emph{MPC-Oracle}~\cite{ZhangRossiEtAl2016b}: we optimize passenger and rebalancing flows using a standard MPC formulation, which assumes perfect foresight of future demand and an estimation of future network conditions (e.g., travel times, prices). In macroscopic settings, where policy decisions do not impact network conditions and future states can be perfectly predicted, this approach serves as an \emph{oracle}, providing an upper bound of performance. In mesoscopic scenarios, policy decisions influence network dynamics, and infrastructure elements such as intersections introduce variability in travel times. As a result, actual travel times may deviate from the forecasts made at the decision-making stage.
    
    \item \emph{MPC-Forecast}: We relax the assumption of perfect foresight in the MPC-Oracle by replacing it with a noisy and unbiased estimate of demand. 
\end{itemize}
We further investigate the impact of key algorithmic design decisions, i.e., neural network architectures and strategies for transfer learning:

\noindent\textbf{Network Architecture}:
\begin{itemize}[leftmargin=*]
    \item \emph{MLP}: we replace the GCN  with a deep (feed-forward) neural network. To enable this model to process the data, we aggregate all node features into a single vector representation.
    \item \emph{CNN}: we substitute the GCN with a deep (convolutional) neural network. To ensure the model can process the data, we structure all node features into a spatial grid.
\end{itemize}

\noindent\textbf{Transfer Learning Strategies}:
\begin{itemize}[leftmargin=*]
    \item \emph{Zero-shot}: The model is deployed directly in the previously unseen test scenario. 
    \item \emph{Fine-tune}: The model is given a limited amount of interaction with the test scenario, to adapt through fine-tuning. 
    \item \emph{Single-city}: the upper-performance limit for transfer methods. It assumes unrestricted environment access, enabling the AMoD controller to be trained until convergence. 
\end{itemize}

\subsection{Macroscopic Scenario}
\label{subsec:macro_results}
\noindent We first present a comprehensive evaluation of our graph-RL framework adopting a \emph{macroscopic} description of traffic dynamics. We begin with an ablation study to assess the role of architectural components and the scalability benefits of our hierarchical policy decomposition. We then examine improvements in the framework’s transfer capabilities, adaptability to network disturbances, and sample efficiency through changes in the learning signal.

\noindent\textbf{Network Architecture.}
\label{subsubsec:network_architecture}
\begin{table}[t]
\caption{Performance on a 16-station network over New York.}
\centering
\begin{tabular}{l | c c c}
     & Profit (k\$) & Served & Rebalancing  \\
     & (\%Dev. MPC) & Demand & Cost (k\$) \\ [0.25ex] 
    \hline
    ED & 30.7 (-14.4\%) & 8.77 & 7.99 \\ [0.2ex]
    MLP & 30.7 (-14.6\%) & 8.77 & 7.92 \\ [0.2ex]
    CNN & 30.4 (-14.8\%) & 8.90 & 8.78 \\ [0.2ex]
    GNN (Graph-RL) & 33.9 (\textbf{-4.3\%}) & 8.77 & 5.04 \\ [0.2ex]
    MPC-Oracle & 35.4 (0\%) & 8.97 & 4.30 \\ [0.2ex]
    \hline
    \end{tabular}%
  \label{tab:nyc_1}%
\end{table}
We evaluate the system performance using a 16-station network over the city of New York (NY). In all experiments, the RL objective corresponds to the operational profit defined in \cref{eq:global_obj}, which is composed of the revenue generated by serving demand minus the rebalancing cost. The results presented in Table \ref{tab:nyc_1} demonstrate that our approach (``Graph-RL") is capable of learning rebalancing policies with near-optimal system performance. Specifically, the performance of Graph-RL deviates from the oracle benchmark only by $1.6\%$. Notably, Graph-RL leverages its learned shared local filter to achieve significant cost savings in rebalancing trips, with reductions of more than $36\%$ compared to learning-based approaches that use different neural architectures.

\noindent\textbf{Computational Analysis.}
\label{subsubsec:computational_analysis}
\begin{figure}[t]
  \centering
  \begin{minipage}[b]{0.49\columnwidth}
    \centering
    \includegraphics[width=\columnwidth]{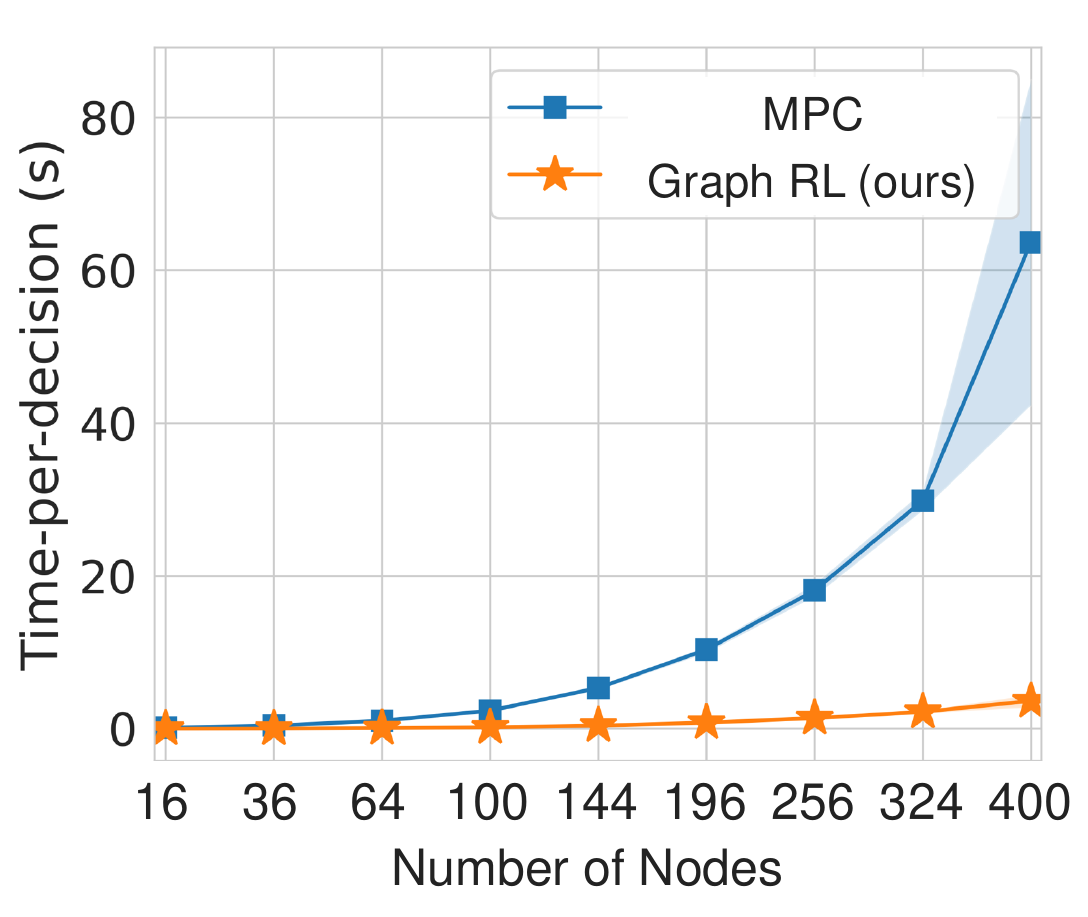}
  \end{minipage}
  \hfill
  \begin{minipage}[b]{0.49\columnwidth}
    \centering
   \includegraphics[width=\columnwidth]{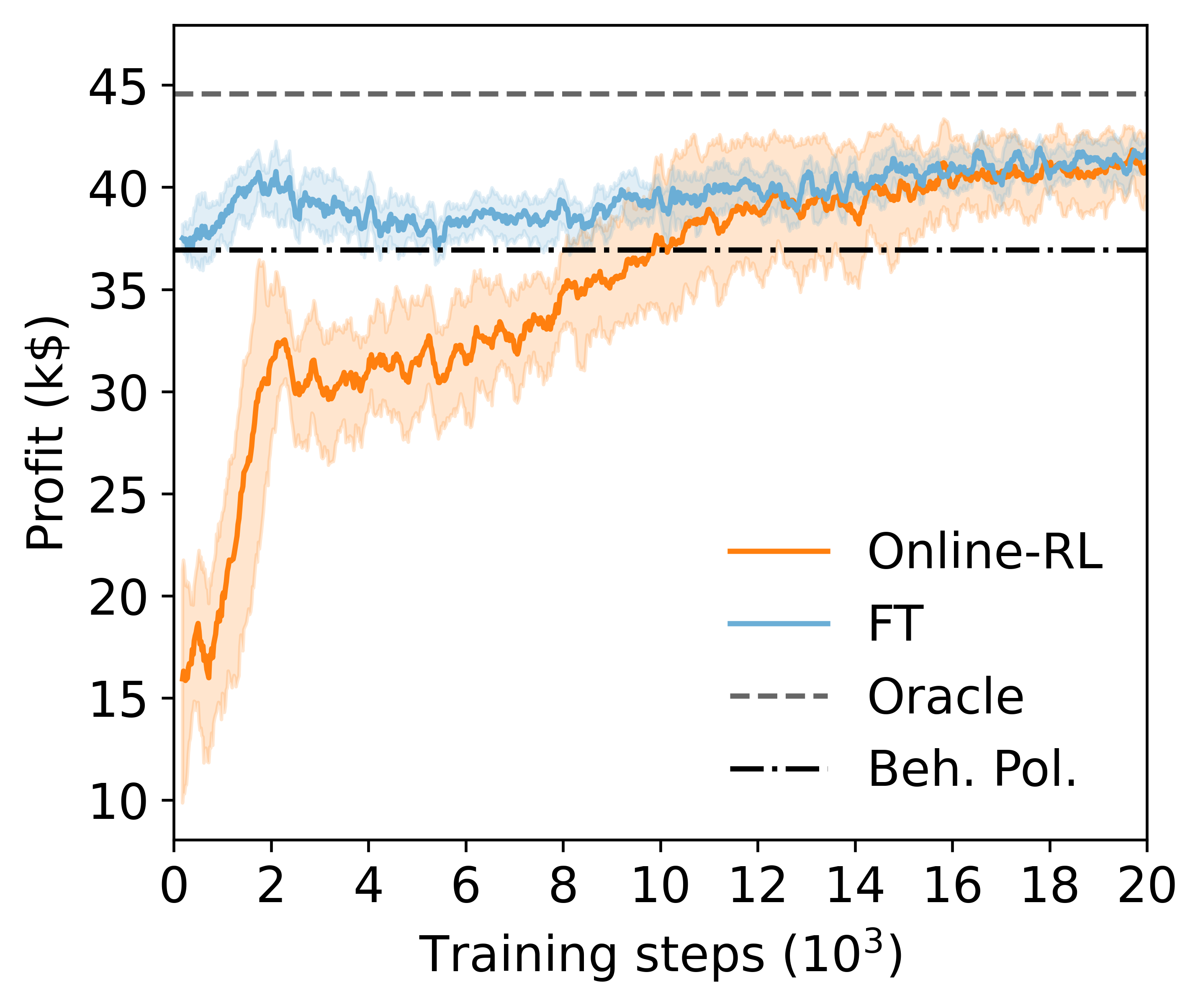}
  \end{minipage}
  \caption{(Left) Comparison of computation times between Graph-RL and MPC. (Right) Fine-tuning performance of pre-trained policy (FT) vs. training from scratch (Online-RL).}
  \label{fig:nyc_granularity}
\end{figure}  
We compare the computational cost of Graph-RL with optimization-based solutions by measuring the time to compute rebalancing decisions across networks of $16$ to $400$ stations. 
Results in Fig. \ref{fig:nyc_granularity} (left) show that Graph-RL efficiently computes actions and scales to large networks by leveraging the efficient inference requirement of neural networks and convex optimization, offering significant computational advantages over control-based approaches.


\noindent\textbf{Inter-city Transferability.}
\label{subsubsec:inter-city_trans}
\begin{table}[t]
    \footnotesize
    \centering
    \caption{Average profit (thousands, \$). \textbf{Black-bold} and \highest{red} highlight best performing model and best performing \emph{adaptive} model, respectively. \textbf{\highest{Red-bold}} is used in case the two coincide.}
    \begin{tabular}{l | c c c}
         & San Francisco & New York Brooklyn & Shenzhen West \\ [0.25ex] 
    \hline
    Zero-shot   & 14.3 ($\pm$ 0.4) & 46.4 ($\pm$ 1.0) & 60.3 ($\pm$ 1.1)\\ [0.2ex] 
    Fine-tune   & 14.6 ($\pm$ 0.3) & 46.0 ($\pm$ 0.7) & 60.9 ($\pm$ 0.8)\\ [0.2ex] 
    Meta-RL     & \textbf{\highest{15.2 ($\pm$ 0.3)}} & \highest{56.2 ($\pm$ 0.7)} & \highest{61.5 ($\pm$ 0.7)}\\
    Single-city & 14.4 ($\pm$ 0.4) & 56.4 ($\pm$ 0.8) & \textbf{63.5 ($\pm$ 0.9)}\\ [0.25ex] 
    MPC-Forecast& 13.9 ($\pm$ 0.5) & \textbf{56.5 ($\pm$ 1.2)} & 60.6 ($\pm$ 1.0)\\ [0.2ex] 
    MPC-Oracle  & 15.9 ($\pm$ 0.4) & 57.2 ($\pm$ 0.8) & 65.2 ($\pm$ 1.0)\\
    \hline
    \end{tabular}
    \label{tab:meta_learning_across_cities}
\end{table}
Current learning-based methods for AMoD control are mostly formulated within a single city paradigm, requiring full re-training for each deployment. To address this, we formalize the multi-city AMoD problem through the lens of meta-RL, enabling controllers to adapt efficiently to new cities with minimal additional interaction. Agents are trained across multiple urban environments and evaluated through simulations on unseen transportation systems of San Francisco, NY Brooklyn, and Shenzhen West. 

\cref{tab:meta_learning_across_cities} shows that meta-RL achieves near-optimal performance, with only a small gap (4.3\%, 1.8\%, and 5.6\%) to the oracle. It creates the most robust policy among the transfer strategies, improving performance by up to 18\% in NY Brooklyn. Crucially, while the single-city approach requires 10,000 training episodes per city, meta-RL achieves similar performance with just 10 episodes—a 1000× improvement in data efficiency. These findings underscore the benefits of incorporating transfer and generalization in AMoD training.

\noindent\textbf{Adaptation to Disturbances.}
\label{subsubsec:online_adaption}
To further assess the generalization performance of our graph meta-RL approach, we evaluate its ability to adapt to real-time disturbances in the NY Brooklyn scenario. While the MPC-Oracle has perfect future knowledge, other methods must react during testing, relying either on current state observations (model-free RL) or imperfect forecasts (MPC-Forecast).

We simulate a sudden demand shift in Forest Hills, Queens. The results, summarized in Table \ref{tab:special_event}, show that meta-RL generates significantly more robust policies than other RL strategies, achieving a 50.3\% improvement over the Single-city approach and outperforming Zero-shot and Fine-tune strategies by 22.1\% and 20.2\%, respectively. Additionally, meta-RL outperforms MPC-Forecast by 4.6\%. These findings highlight the limitations of training RL agents solely in single-city scenarios, as such policies may be highly sensitive to changes in distributions.

\begin{table}[t]
\footnotesize
\centering
\caption{Special event scenario in New York Brooklyn.}
\begin{tabular}{l | c c c}
    & Profit (k\$)  & Served Demand& Reb. Cost (k\$)\\
    \hline 
    Zero-Shot & 38.8 ($\pm$ 0.95) & 1.64 & 7.89\\ [0.1ex]
    Fine-tune & 39.7 ($\pm$ 1.59) & 1.63 & 8.15 \\[0.1ex]
    Meta-RL & \highest{49.8 ($\pm$ 1.35)} & 2.56 & 15.64  \\[0.1ex]
    Single-city & 24.7 ($\pm$ 5.00) & 1.71 & 11.05  \\[0.1ex]
    MPC-Forecast & 47.5 ($\pm$ 2.39) & 2.98 & 16.22 \\[0.1ex]
    MPC-Oracle & 57.5 ($\pm$ 0.81) & 3.11 & 7.73 \\[0.1ex]
    \hline
    \end{tabular}%
\label{tab:special_event}%
\end{table}

\noindent\textbf{Sample Efficiency.}
\label{subsubsec:offline_rl}
Online RL approaches assume access to a reliable mobility simulator and require extensive online interactions. 
Offline RL and BC offer a promising alternative by learning directly from past data, reducing dependence on costly online interaction. To evaluate whether a policy trained on a static dataset can learn effective rebalancing strategies, we generate datasets using three behavior policies: informed rebalancing (INF) \cite{WallarEtAl2018}, dynamic trip-vehicle assignment (DTV) \cite{Alonso-MoraEtAl2017}, a demand-proportional heuristic (PROP). Table \ref{tab:mod} compares online RL (SAC), which freely interacts with the environment until convergence, to offline methods (BC, IQL, CQL), which learn from a static dataset \textit{without any environment interactions}. 
The results highlight how BC approaches are limited by the dataset performance, while offline RL algorithms (IQL, CQL) can successfully outperform the behavior policy. 
Finally, we assess the weakest offline policy (trained on PROP) during online fine-tuning. Figure~\ref{fig:nyc_granularity} (right) shows that the offline RL policy reliably improves upon its initial performance, while \textit{consistently remaining above the performance of the behavior policy}. In contrast, training a policy from scratch requires over 12,000 online steps to match the performance of the pre-trained agent, incurring significant cost within our simulations---$\$98,265$ in lost revenue and $\$1,029,162$ in excess rebalancing costs---clearly quantifying the benefits of offline learning.

\begin{table}[t]
\footnotesize
\centering
\caption{Profit (thousands, \$) comparing online (SAC) and offline (BC, IQL, CQL) algorithms in NY Brooklyn.}
\begin{tabular}{l l | c c c l}
    Dataset & Beh. Pol. & SAC & BC & IQL & CQL \\
    \hline 
    INF & 56.9 & 56.1 $\pm$1.1 & 56.5 $\pm$1.3 & \textbf{56.8} $\pm$1.6 & 53.8 $\pm$1.0 \\
    DTV & 51.7 & 56.1 $\pm$1.1 & 51.6 $\pm$1.4 & \textbf{52.7} $\pm$1.6 & 48.3 $\pm$1.3 \\
    PROP & 49.6  & 56.1 $\pm$1.1 & 49.6 $\pm$1.4 & 49.6 $\pm$1.3 & \textbf{50.9} $\pm$1.4 \\
    \hline
    \end{tabular}%
\label{tab:mod}%
\end{table}

\subsection{Mesoscopic Scenario}
\label{subsec:meso_results}
\noindent In this section, we evaluate the performance of the proposed hierarchical policy leveraging a high-fidelity \emph{mesoscopic} traffic simulator of the city of Luxembourg~\cite{Codeca2017} with a SAC agent~\cite{Haarnoja2018}. 
The city network is aggregated into 10 regions (2 km$^2$ each) and analyzed during the afternoon peak (4 pm – 6 pm) using both operator- and network-oriented metrics.
We then explore biasing policy training to assess how the design of the reward function can elicit desired behavior from learning-based policies. Next, we assess the policy's generalization capabilities across simulator fidelities, examining its transferability from low- to high-fidelity environments.


Performance is measured using two groups of metrics: profit-related metrics, including operator profit, revenue, and rebalancing costs, and network-related metrics, i.e., average passenger waiting time (network-wide and per region) and average fleet utilization (UF).
Additional details and experiments testing policy generalization can be found in \cref{app:meso_metrics}.

\noindent\textbf{Coordination During Afternoon Peak Time.}
\label{subsubsec:res_base_scenario}
In our first simulation experiment, we evaluate the system's ability to coordinate an AMoD fleet during the afternoon traffic peak.

\begin{table}[t]
\footnotesize
\centering
\caption{System Performance on Luxembourg mesoscopic simulation during the afternoon peak. Operator metrics (left), and network metrics (right).}
\begin{tabular}{l | c c | c c}
         & Profit (k\$)   & Reb. & Avg. Wait. & Avg.\\
         & (\% Dev. MPC) &  Cost (k\$) & Time (s) & UF (\%)\\
    \hline
        No Reb. & 34.7 (-7.8\%) & 0 & 269 & 57 \\ [0.2ex]
        ED & 36.0 (-4.2\%) & 1.90 & 143 & 80 \\ [0.2ex]
        P1 & 35.3 (-6.0\%) & 1.39 & 238 & 76 \\ [0.2ex]
        Graph-RL & 36.7 (\textbf{-2.3\%}) & 1.35 & 133 & 75 \\ [0.2ex]
        MPC-Forecast & 37.4 (-0.5\%) & 0.44 & 281 & 65\\ [0.2ex]
        MPC-Oracle & 37.6 \phantom{00}(0\%) & 0.56 & 165 & 69\\ [0.2ex]
    \hline
    \end{tabular}%
  \label{tab:lux10_aggr}%
\end{table}

\noindent\textit{Operator metrics.}
Results in \cref{tab:lux10_aggr} demonstrate that Graph-RL effectively learns rebalancing policies that achieve near-optimal performance, even in a high-fidelity traffic environment.
Specifically, Graph-RL delivers a profit within 2.3\% of the MPC oracle while outperforming all heuristic methods.
Notably, it serves the highest demand, matching the performance of the ED and MPC policies while reducing rebalancing costs by 29\% compared to ED.
This reflects a more strategic use of idle vehicles.
In contrast, the MPC strategy minimizes rebalancing costs by avoiding less profitable trips, resulting in a more static policy and slightly lower revenue.

\begin{figure}
\centering
    \subfloat[No Reb.]{
      \includegraphics[width=.5\columnwidth]{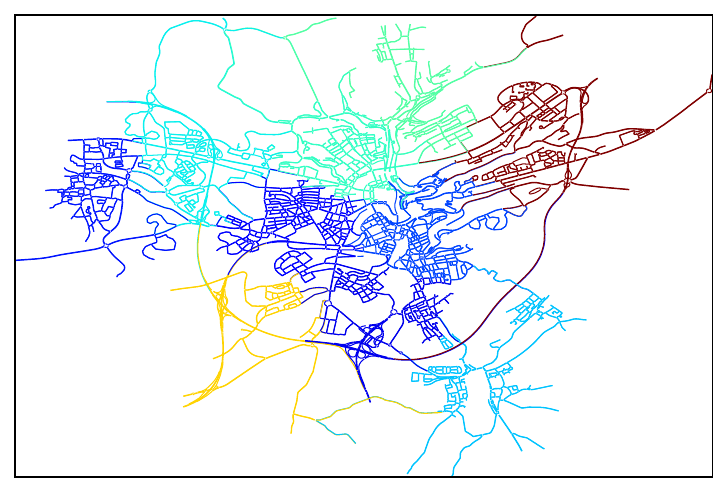}
    }
    \subfloat[ED]{
      \includegraphics[width=.5\columnwidth]{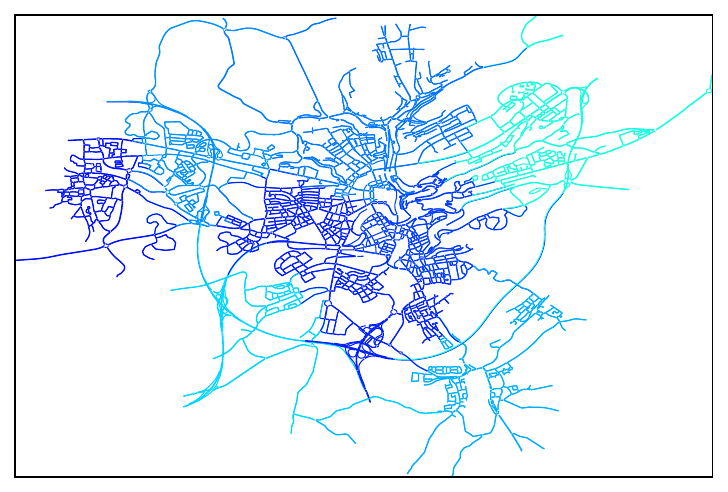}
    }
    \hspace{0mm}
    \subfloat[Graph-RL]{
      \includegraphics[width=.5\columnwidth]{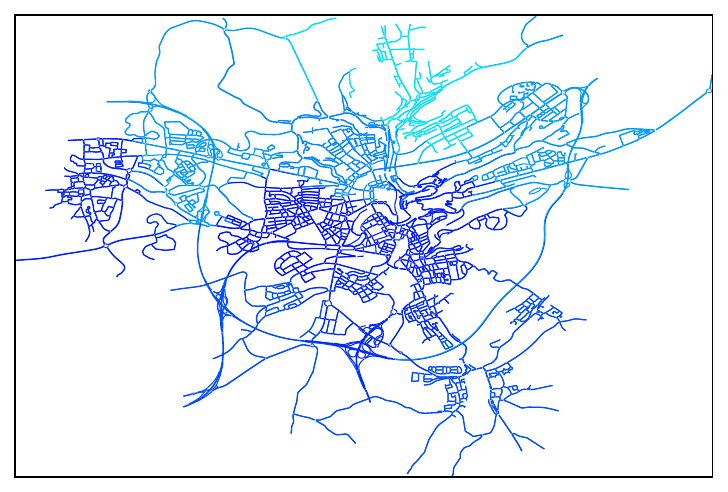}
    }
    \subfloat[MPC-oracle]{%
    \raisebox{-0.8mm}{%
    \hspace{0.2mm}
      \includegraphics[width=.5\columnwidth]{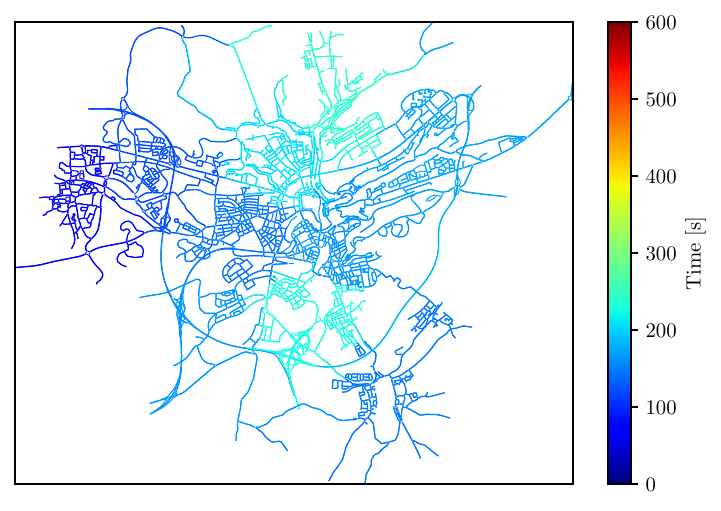}}
    }
\caption{Average waiting time per passengers distribution across the city, among the policies considered in the study.}
\label{fig:waiting_time_distribution}
\end{figure}

\noindent\textit{Network metrics.}
From a network perspective, Graph-RL demonstrates superior performance in enhancing customer satisfaction compared to other approaches. 
It achieves the lowest average passenger waiting time, driven by its active fleet utilization of 75\%, as shown in \cref{tab:lux10_aggr}.
In contrast, the MPC policy adopts a more conservative strategy, utilizing only $69\%$ of the fleet on average, which leads to longer waiting times despite its cost efficiency.
While the ED policy achieves the highest fleet utilization (80\%), its reactive approach results in higher rebalancing costs and inefficiencies, as unnecessary rebalancing trips reduce vehicle availability for passenger matching.
Consequently, the ED policy delivers an average waiting time that is 7.52\% higher than with Graph-RL.
Analyzing the spatial distribution of waiting times in Figure \ref{fig:waiting_time_distribution}, reveals that Graph-RL, MPC, and ED policies provide balanced service coverage across the city. These policies efficiently distribute resources, ensuring passengers are promptly served regardless of location. In contrast, the No Rebalancing policy exhibits significant passenger accumulation in rural areas. This disparity explains its higher average waiting times despite similar average fleet utilization to Graph-RL. 

\begin{figure}[t]
      \centering
     \includegraphics[width=0.95\columnwidth]{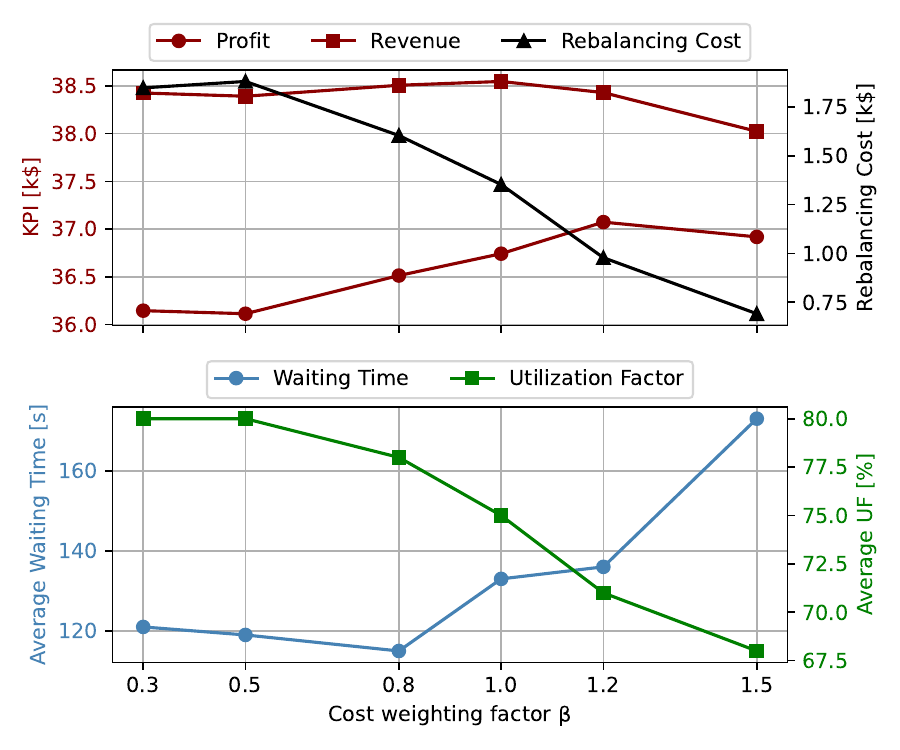}
      \caption{Operator and network performance across reward structures. (Top) Profit, revenue, and rebalancing cost. (Bottom) Average waiting time and fleet utilization factor.}
\label{fig:beta_sweep}
\end{figure}

\noindent\textbf{Reward Design.}
\label{subsubsec:res_beta_sweep}
We extend our analysis of the Graph-RL policy by examining how changes in the cost structure for vehicle movement affect both operator and network performance.
Specifically, we introduce a penalization term proportional to travel time,~$c_{ij} =\beta\tau_{i,j}$, which directly influences the reward signal.
An evaluation of varying~$\beta$ assesses its influence on policy behavior, as shown in \cref{fig:beta_sweep}.
For low penalization values ($\beta \leq 0.5$), the policy maintains stable behavior, yielding consistent performance from both operator and network perspectives.
However, as~$\beta$ increases beyond 0.5, the policy exhibits increasingly static behavior.
On the operator side, more penalization encourages a cost-conscious strategy, significantly reducing rebalancing costs while maintaining stable revenue and improving profitability.
Conversely, from a network perspective, reduced fleet activity leads to longer average waiting times and decreased fleet utilization.
These results underscore the trade-offs of increasing vehicle movement penalization: while lower costs boost operational efficiency, they may negatively affect passenger experience and fleet performance.
Therefore, careful calibration of the reward signal is essential to develop profitable policies that also sustain high levels of customer satisfaction.



\noindent\textbf{Transfer Performance Across Simulator Fidelities.}
\label{subsubsec:meso_transfer}
We evaluate the transferability of the Graph-RL policy trained in a macroscopic environment to a higher-fidelity mesoscopic traffic simulation \emph{without additional training}.
This approach aims to determine whether training costs can be reduced, as macroscopic training requires only 17 hours compared to 5 days for mesoscopic training\footnote{All methods used the same computational CPU resources, namely a AMD Ryzen 9 7950X3D (16-Core, 32-Thread, 144MB Cache, 4.2 GHz base).}.
\cref{tab:gen_macro_to_meso} presents a comparison of the zero-shot performance of the Graph-RL agent, trained during the afternoon peak in the macroscopic setting (denoted as zero-shot), against baseline policies and a Graph-RL model trained directly in the mesoscopic setting.
The results highlight the potential of leveraging low-fidelity simulations to enhance computational efficiency while maintaining strong performance. 

\begin{table}[tb]
\caption{Transfer performance across simulator fidelity on a Luxembourg mesoscopic simulation (4–6 pm).}
\centering
\footnotesize
\begin{tabular}{l | c c c}
         & Profit  & Revenue  & Reb.\\
         & (k\$) & (k\$) &  Cost (k\$)\\
    \hline
        No Reb. & 34.7 (-7.8\%) & 34.7 & 0 \\ [0.2ex]
        ED & 36.0 (-4.2\%) & 38.3 & 1.90 \\ [0.2ex]
        P1 & 35.3 (-6.0\%) & 37.1 & 1.39 \\ [0.2ex]
        Graph-RL (zero-shot) & 36.4 (\textbf{-3.2\%}) & 38.5 & 1.67 \\ [0.2ex]
        Graph-RL (meso) & 36.7 (\textbf{-2.3\%}) & 38.5 & 1.35 \\ [0.2ex]
        MPC-oracle & 37.6 (0\%) & 38.2 & 0.56 \\ [0.2ex]
    \hline
    \end{tabular}%
  \label{tab:gen_macro_to_meso}%
\end{table}
The results demonstrate excellent cross-fidelity generalization with the zero-shot policy achieving a 1.1\% profit reduction compared to the performance of the Graph-RL agent retrained in the mesoscopic environment. 
Notably, the zero-shot policy also outperforms all heuristic baselines in the mesoscopic setting, showcasing strong robustness.
These findings indicate that training in a lower-fidelity macroscopic environment effectively captures high-level system dynamics that translate well to higher-fidelity simulations.
Cross-fidelity generalization offers significant benefits, including reduced computational overhead and shorter training times.
Furthermore, this approach underscores the potential of leveraging the efficiency of macroscopic simulations for initial training phases.
Overall, the results reinforce the Graph-RL framework's ability to learn a generalized understanding of system dynamics across spatial, temporal, and fidelity domains, supporting scalable and resource-efficient policy development for complex transportation systems.

\section{Conclusion}
\noindent We presented a novel framework for robo-taxi fleet coordination that integrates optimization, learning, and graph representation learning.
Our approach demonstrates robust performance across diverse urban scenarios and simulation fidelities, achieving near-optimal results with superior computational efficiency and transferability.
The proposed hierarchical policy decomposition, combined with meta-RL, improves adaptability to dynamic environments and reduces the need for retraining.
Furthermore, the release of an open-source codebase, including benchmarks and simulators, aims to support future research and establish a standardized evaluation process for AMoD systems.
Our findings highlight the potential of advanced learning-based methods to transform autonomous fleet management, paving the way for efficient mobility solutions.

Future work will explore extending these methods to more complex multi-agent scenarios, integrating real-time data streams (e.g., disruptions and demand adaptations), and optimizing for additional objectives, such as energy efficiency (e.g., for electric vehicles) and equity in service distribution.
\label{sec:conclusion}

\bibliographystyle{unsrt} 
\bibliography{main, ASL_Bib}

\appendix

\subsection{Mesoscopic Scenario Metrics}
\label{app:meso_metrics}
\noindent We consider two groups of key performance indicators (KPIs): operator and network/customer metrics. 
The operator KPIs consider the \emph{operational profit}, which is defined as the revenue from \emph{served demand} minus the \emph{rebalancing cost}.
Let \(M_{\text{reb}}^t\) and \(M_{\text{match}}^t\) denote the number of vehicles used for rebalancing and serving passengers at time \(t\), with total fleet size \(M_{\text{tot}}\). We consider the \emph{fleet utilization factor} $UF = \frac{M_{\text{reb}}^t + M_{\text{match}}^t}{M_{\text{tot}}}$. 
From the customer point of view, in a scenario with a total passenger demand of \(D=\sum_{t=1}^{T}\sum_{i,j}d_{i,j}^t\), and a $i$th regional demand of \(D_i=\sum_{t=1}^{T}\sum_{j}d_{i,j}^t\), we look at the \emph{overall average waiting time} $t_{\text{wt,avg}} = \frac{\sum_{k=1}^D t_{\text{wt},k}}{D}$, and the \emph{regional average waiting time} $t_{\text{wt,avg},i} = \frac{\sum_{k=1}^{D_i} t_{\text{wt},k}}{D_i}$.

\subsection{Transfer Performance on the Mesoscopic Scenario}
\label{app:meso_gen}

\noindent We assess the transfer performance of a \emph{baseline policy} (trained during the 4–6 pm peak, on a 10-region network). 

\begin{figure}[t]
      \centering     \includegraphics[width=1\columnwidth]{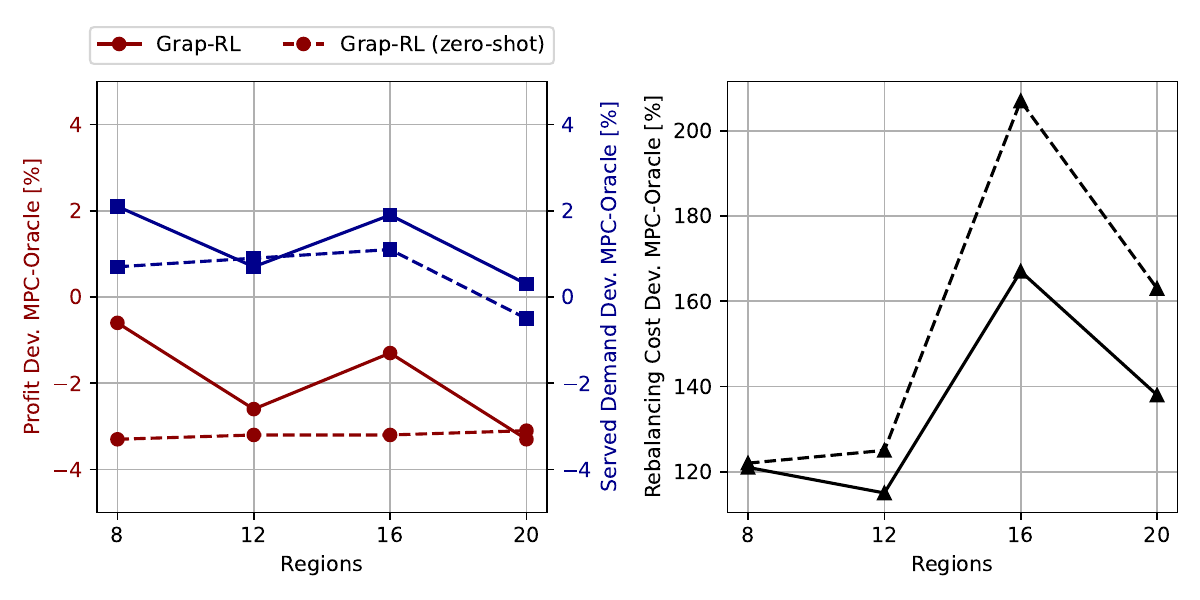}
      \caption{Comparison of operator KPIs between a fully retrained Graph-RL policy and its zero-shot performance across different network granularity, measured as deviations from Oracle.}
\label{fig:gen_region}
\end{figure}

\noindent\textbf{Across Spatial Aggregation.}
Figure \ref{fig:gen_region} compares the performance of the retrained policy and the zero-shot performance of the baseline policy. 
The retrained policies perform consistently across different spatial aggregation levels, with profit deviations from the MPC-oracle ranging between $0.6\%$ and $3.3\%$. 
Meanwhile, the zero-shot baseline exhibits notable spatial generalization, showing only a moderate performance drop relative to the retrained policy, with profit deviations from $3.1\%$ to $4.4\%$. 
These results confirm Graph-RL’s adaptability to varying region configurations in a high-fidelity scenario.

\noindent\textbf{Across Time of the Day.}
We consider a scenario during \emph{night} (12-02 am), in which the system experiences low demand, a small fleet, and almost free-flow traffic; and during the \emph{morning rush}, which presents similar conditions to the baseline (afternoon) peak, but with a more congested network.

\begin{figure}[t]
      \centering
     \includegraphics[width=1\columnwidth]{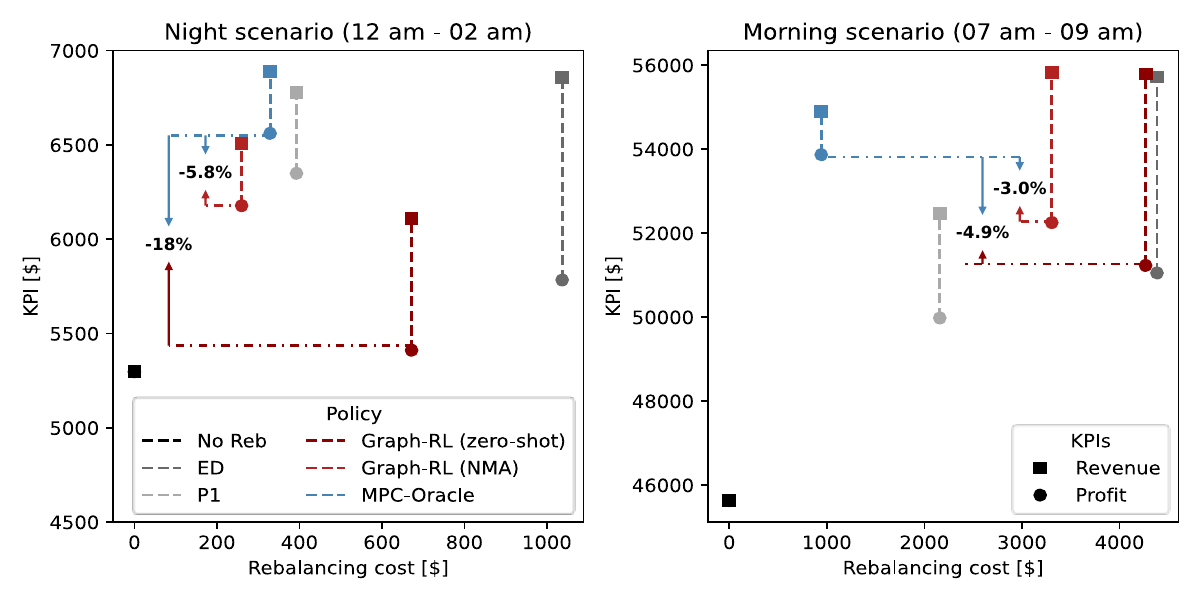}
      \caption{Revenue (square) and profit (circle) vs. rebalancing cost during night (left) and morning rush (right). 
      Displayed are the percentage deviations in profit relative to the oracle.}
\label{fig:gen_time}
\end{figure}

\noindent Figure \ref{fig:gen_time} compares the baseline policies to an RL agent trained on a combined dataset spanning night, morning, and afternoon (NMA). 
\noindent During the night, the Graph-RL agent’s profit drops up to $–18\%$ compared to the MPC-oracle. 
In contrast, during the morning rush, the network conditions are more aligned with the baseline afternoon scenario, resulting in a lower profit gap of $–4.9\%$. 
Extending the training dataset leads to improvements in generalization, as the NMA policy achieves profit deviations of $–5.8\%$ (night) and $–3.0\%$ (morning). 

\vspace{-2em}
\begin{IEEEbiographynophoto}
{Luigi Tresca}
is a Ph.D. student at Politecnico di Torino in Italy, where his work is focused on the integration of learning-based models and vehicle connectivity information into the energy management system of hybrid electric vehicles. During his visiting period at the Autonomous Systems Lab at Stanford University, he worked on the integration of realistic traffic simulators for training and testing learning-based controllers for Autonomous Mobility on Demand systems.
\end{IEEEbiographynophoto}
\vspace{-3em}
\begin{IEEEbiographynophoto}
{Carolin Schmidt} is a Postdoctoral Scholar at the Technical University of Denmark (DTU), where her research focuses on learning-based control for autonomous mobility systems. As part of her PhD at DTU, she was involved in an EU-wide project on the real-world deployment of automated shuttles. As a visiting researcher at the Autonomous Systems Lab at Stanford University, she worked on integrating deep RL with optimization and control methods to create efficient and robust RL, contributing to applications in transportation and robotics. 
\end{IEEEbiographynophoto}
\vspace{-3em}
\begin{IEEEbiographynophoto}
{James Harrison} is a research scientist at Google DeepMind, where his work focuses on improving decision-making with learning. In particular, his work includes learning for optimization; scalable methods for robustness and uncertainty in large neural networks; RL and optimization for industrial engineering and operations research applications; and learning, reasoning, and task planning for robots. He received his Ph.D. in mechanical engineering from Stanford University in 2021, where his research was on safe, continual learning for autonomous robots. 
\end{IEEEbiographynophoto}
\vspace{-3em}
\begin{IEEEbiographynophoto}
{Filipe Rodrigues} is associate professor at the Technical University of Denmark (DTU) in the Machine Learning for Smart Mobility lab, where his research is primarily focused on machine learning models for understanding and optimizing urban mobility and human behavior. Previously, he was a H.C. Ørsted / Marie-Skłodowska Curie Actions postdoctoral fellow, also at DTU, working on spatio-temporal models of mobility demand with emphasis on modeling uncertainty and the effect of special events. 
His research interests span machine learning, RL, intelligent transportation systems, and urban mobility. 
\end{IEEEbiographynophoto}
\vspace{-3em}
\begin{IEEEbiographynophoto}
{Gioele Zardini}
is the Rudge (1948) and Nancy Allen Assistant Professor at Massachusetts Institute of Technology. He is a PI in the Laboratory for Information and Decision Systems (LIDS) and an affiliate faculty with the Institute for Data, Systems and Society (IDSS). He received his doctoral degree in 2024 from ETH Zurich, and holds both a B.Sc. and a M.Sc. in Mechanical Engineering and Robotics, Systems and control from ETH Zurich. Before joining MIT, he was a Postdoctoral Scholar at Stanford University and held various visiting positions at nuTonomy Singapore (then Aptiv, now Motional), Stanford, and MIT. He is the recipient of an award (keynote talk) at the 2021 Applied Category Theory Conference, the Best Paper Award (1st Place) at the 2021 24th IEEE International Conference on Intelligent Transportation Systems, and Amazon and DoE awards.
\end{IEEEbiographynophoto}
\vspace{-3em}

\begin{IEEEbiographynophoto}{Daniele Gammelli} is a Postdoctoral Scholar in the Autonomous Systems Lab at Stanford University. 
He received his Ph.D. in Machine Learning and Mathematical Optimization at the Technical University of Denmark. 
His research focuses on developing learning-based solutions that enable the deployment of future autonomous systems in complex environments, with an emphasis on large-scale robotic networks, aerospace systems, and future mobility systems.
Dr. Gammelli has been making research contributions in fundamental AI research, robotics, and its applications to network optimization, mobility systems, and autonomous aerospace.
\end{IEEEbiographynophoto}
\vspace{-3em}
\begin{IEEEbiographynophoto}
{Marco Pavone} is an Associate Professor of Aeronautics and Astronautics at Stanford University, where he directs the Autonomous Systems Laboratory and the Center for Automotive Research at Stanford. He is also a Distinguished Research Scientist at NVIDIA, where he leads autonomous vehicle research. Before joining Stanford, he was a Research Technologist within the Robotics Section at the NASA Jet Propulsion Laboratory. He received a Ph.D. degree in Aeronautics and Astronautics from the Massachusetts Institute of Technology in 2010. His main research interests are in the development of methodologies for the analysis, design, and control of autonomous systems, with an emphasis on self-driving cars, autonomous aerospace vehicles, and future mobility systems. He is a recipient of a number of awards, including a Presidential Early Career Award for Scientists and Engineers from President Barack Obama.
\end{IEEEbiographynophoto}
\end{document}

%% file: math_commands.tex
\usepackage{amsmath,amsfonts,bm}

\newcommand{\numvehicles}{M}
\newcommand{\graph}{\gG}
\newcommand{\nodes}{\gV}
\newcommand{\edges}{\gE}
\newcommand{\edge}{e}
\newcommand{\numregions}{N_{\gV}}
\newcommand{\horizon}{\gT}
\newcommand{\ttime}{\tau}
\newcommand{\tcost}{c}
\newcommand{\demand}{d}
\newcommand{\price}{p}
\newcommand{\maxwait}{\tau_{\mathrm{max}}}
\newcommand{\bx}{\mathbf{x}}
\newcommand{\paxflow}{x}
\newcommand{\rebflow}{y}
\newcommand{\pflatopt}{\tilde \pi^*}

\newcommand{\pflatspace}{\tilde \Pi}

\newcommand{\st}{\mathbf{s}}
\newcommand{\ac}{\mathbf{a}}

\newcommand{\inventory}{u}

\def\eqref#1{equation~\ref{#1}}

\def\1{\bm{1}}

\DeclareMathAlphabet{\mathsfit}{\encodingdefault}{\sfdefault}{m}{sl}
\SetMathAlphabet{\mathsfit}{bold}{\encodingdefault}{\sfdefault}{bx}{n}

\def\gE{{\mathcal{E}}}

\def\gG{{\mathcal{G}}}

\def\gT{{\mathcal{T}}}

\def\gV{{\mathcal{V}}}

